\documentclass[letterpaper, 10pt]{ieeeconf}
\IEEEoverridecommandlockouts

\usepackage{cite}
\usepackage{amsmath,amssymb,amsfonts}
\usepackage{bm}
\usepackage{algorithmic}
\usepackage{graphicx}
\usepackage{textcomp}
\usepackage{xcolor,color}
\usepackage[ruled,vlined]{algorithm2e}
\usepackage[top=0.833in, left=0.667in, right=0.667in, bottom=0.597in, includefoot]{geometry}
\usepackage{hyperref}
\usepackage{tikz}
\usepackage{gensymb}
\usepackage{arydshln}
\usepackage{bbold}

\usetikzlibrary{positioning}

\usepackage{amsthm}
    
\makeatletter
\let\MYcaption\@makecaption
\makeatother 
\usepackage[font=footnotesize]{subcaption}
\makeatletter
\let\@makecaption\MYcaption
\makeatother

\newcommand\undermat[2]{%
  \makebox[0pt][l]{$\smash{\underbrace{\phantom{%
    \begin{matrix}#2\end{matrix}}}_{\text{$#1$}}}$}#2}

\newcommand{\hl}[1]{#1}

\newcommand{\q}{\bm{q}}
\renewcommand{\v}{\bm{v}}
\newcommand{\vd}{\dot{\v}}
\newcommand{\btau}{\bm{\tau}}

\newcommand{\M}{\bm{M}}
\newcommand{\C}{\bm{C}}
\renewcommand{\S}{\bm{S}}
\newcommand{\J}{\bm{J}}

\newcommand{\f}{\bm{f}}

\newcommand{\Q}{\bm{Q}}
\newcommand{\R}{\bm{R}}

\newcommand{\x}{\bm{x}}

\renewcommand{\u}{\bm{u}}

\newcommand{\p}{\bm{p}}
\newcommand{\g}{\bm{g}}

\newcommand{\K}{\bm{K}}

\begin{document}

\title{Contact-Implicit Trajectory Optimization with Hydroelastic Contact and iLQR}

\author{Vince Kurtz and Hai Lin
\thanks{The authors are with the Departments of Electrical Engineering, University of Notre Dame, Notre Dame, IN, 46556 USA. \texttt{\{vkurtz,hlin1\}@nd.edu}}
\thanks{This work was supported by NSF Grants CNS-1830335, IIS-2007949. The first author is supported by a Dolores Z. Liebmann fellowship.}
}

\maketitle
\thispagestyle{empty}

\begin{abstract}
    Contact-implicit trajectory optimization offers an appealing method of automatically generating complex and contact-rich behaviors for robot manipulation and locomotion. The scalability of such techniques has been limited, however, by the challenge of ensuring both numerical reliability and physical realism. In this paper, we present preliminary results suggesting that the Iterative Linear Quadratic Regulator (iLQR) algorithm together with the recently proposed pressure-field-based hydroelastic contact model enables reliable and physically realistic trajectory optimization through contact. We use this approach to synthesize contact-rich behaviors like quadruped locomotion and whole-arm manipulation. Furthermore, open-loop playback on a Kinova Gen3 robot arm demonstrates the physical accuracy of the whole-arm manipulation trajectories. Code is available at \url{https://bit.ly/ilqr_hc} and videos can be found at \url{https://youtu.be/IqxJKbM8_ms}.
\end{abstract}

\section{Introduction and Related Work}\label{sec:intro}

Many important tasks involve making and breaking contact. As humans, we make contact with the environment to move ourselves through it (locomotion), as well as to move objects relative to ourselves (manipulation). Earlier work on robot locomotion and manipulation focused primarily on predefined contact sequences from reduced-order models (for locomotion) \cite{wieber2016modeling} or grasping heuristics (for manipulation) \cite{mason2018toward}.

More recently, there has been a trend toward discovering contact sequences automatically using trajectory optimization. Contact-implicit trajectory optimization has been used to generate a wide variety of behaviors for manipulation and locomotion \cite{tassa2012synthesis,posa2014direct,manchester2020variational,aydinoglu2021real,patel2019contact,onol2020tuning,cleac2021fast}. Furthermore, this approach can not only synthesize but also stabilize contact-rich trajectories, either through Model Predictive Control (MPC) or with local feedback generated by optimization algorithms like iLQR \cite{li2004iterative}.

The major promise of contact-implicit trajectory optimization---reliable automated generation of contact-rich behaviors---has yet to be realized, however. This is largely due to two conflicting challenges: numerical reliability and physical realism. Complementarity-based rigid contact models \cite{anitescu1997formulating} are physically accurate but lead to non-smooth system dynamics, which present a challenge for gradient-based optimization. While there has been significant recent progress in developing numerically stable optimization algorithms for such rigid contact, existing techniques are typically limited to simple (often linear) dynamics \cite{aydinoglu2021real} or relatively simple contact configurations \cite{sleiman2019contact,carius2018trajectory}. 

On the other hand, compliant contact models such as that used in the MuJoCo simulator \cite{todorov2014convex} are amenable to gradient-based trajectory optimization techniques like iLQR, and have resulted in many impressive simulation demonstrations involving incredibly complex contact configurations and high degree-of-freedom systems \cite{tassa2012synthesis,patel2019contact,onol2020tuning,chatzinikolaidis2021trajectory}. But these contact models include non-physical force-at-a-distance and ``gliding'' artifacts, often resulting in trajectories that are difficult to reproduce on hardware. 

\begin{figure}
    \begin{subfigure}{0.49\linewidth}
        \centering
        \includegraphics[width=\linewidth]{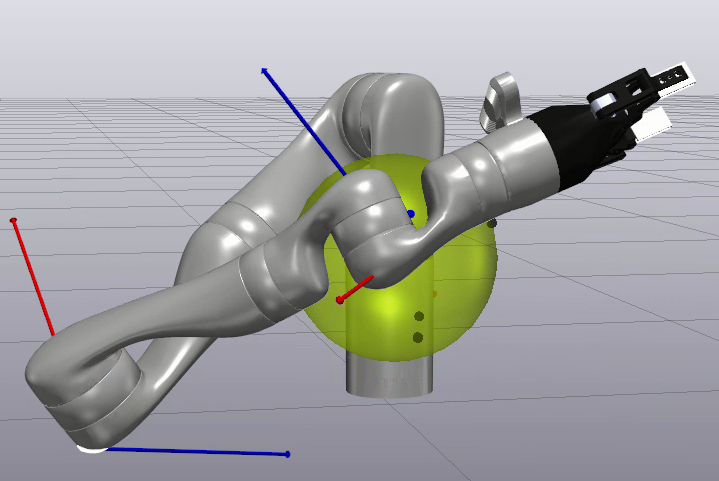}
        \caption{}
        \label{fig:lift_simulation}
    \end{subfigure}
    \begin{subfigure}{0.49\linewidth}
        \centering
        \includegraphics[width=\linewidth]{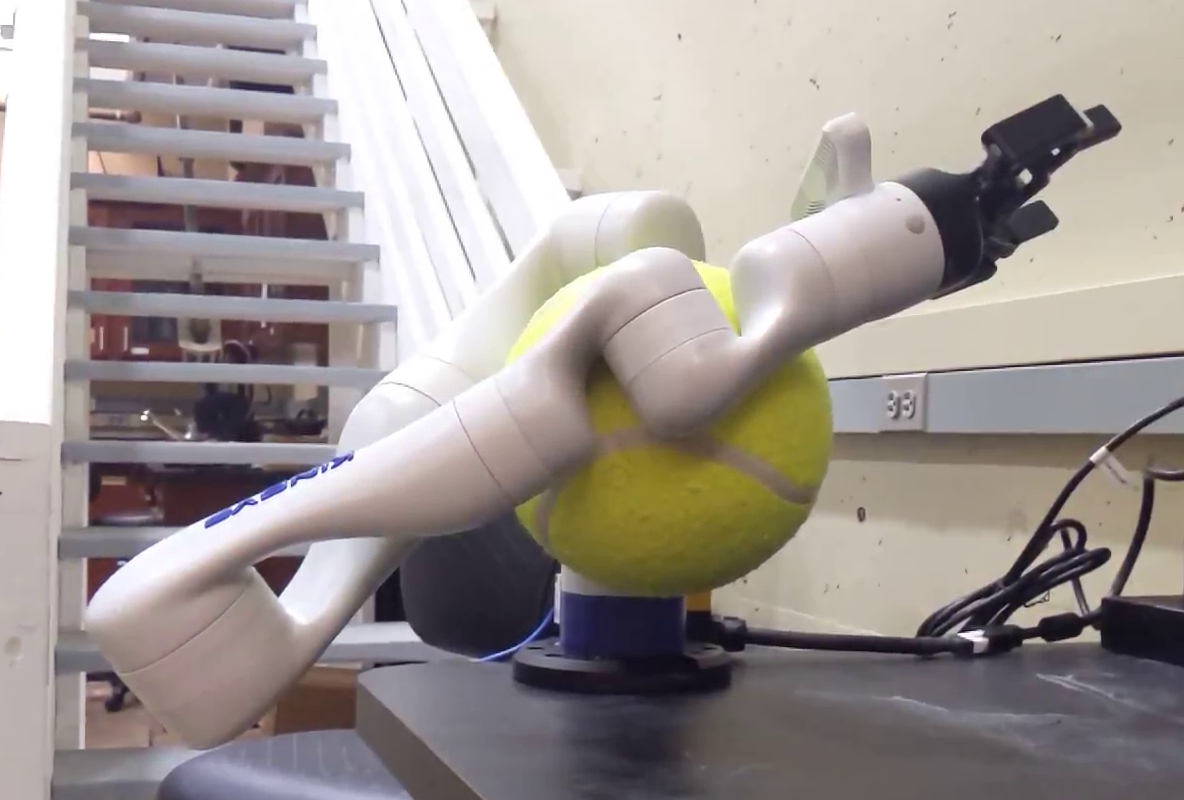}
        \caption{}
        \label{fig:lift_hardware}
    \end{subfigure}
    \begin{subfigure}{\linewidth}
        \centering
        \includegraphics[width=0.7\linewidth]{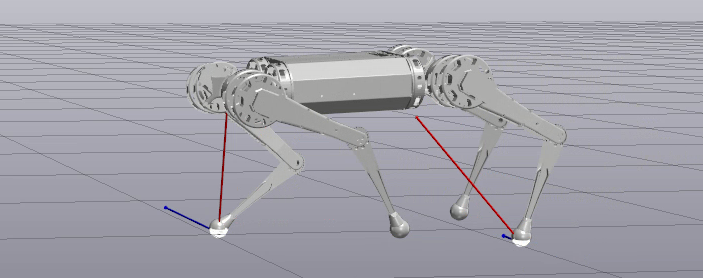}
        \caption{}
        \label{fig:quad_gait}
    \end{subfigure}
    \centering
    \caption{We show that iLQR over hydoelastic contact \cite{elandt2019pressure} enables reliable synthesis of contact-rich behaviors including whole-arm manipulation (\subref{fig:lift_simulation}) and quadruped gait generation (\subref{fig:quad_gait}). Optimal trajectories for whole-arm manipulation could be directly applied on hardware, with little sim-to-real gap (\subref{fig:lift_hardware}).}. 
    \label{fig:front_page}
\end{figure}

In this paper, we consider contact-implicit trajectory optimization over the recently proposed pressure-field-based hydroelastic contact model \cite{elandt2019pressure,masterjohn2021discrete}. Hydroelastic contact allows some interpenetration between nominally rigid objects, and computes contact forces based on a surface integral rather than a single point of maximum penetration. In addition to being physically realistic---this contact model can accurately model interactions between complex geometries like the arm and ball shown in Figure~\ref{fig:hydroelastic}---dynamics gradients can be computed with automatic differentiation. 

We show that iLQR over hydroelastic contact can generate a variety of complex contact-rich behaviors, including whole-arm manipulation for a Kinova Gen3 robot and gait discovery for a Mini Cheetah quadruped. While our simple Python implementation of iLQR was far too slow for real-time MPC, open-loop playback of optimal trajectories on a Kinova Gen3 robot arm resulted in behavior that closely matched the simulation. These preliminary results support the physical realism of the hydroelastic contact model \cite{elandt2019pressure}, and suggest that iLQR over hydroelastic contact is a promising basis for contact-implicit trajectory optimization.

The remainder of this paper is organized as follows: a problem formulation is presented in Section~\ref{sec:problem}, a basic overview of hydroelastic contact and iLQR are presented in Section~\ref{sec:background}, we describe simulation and hardware experiments in Sections \ref{sec:simulation} and \ref{sec:hardware}, discuss advantages and limitations in Section~\ref{sec:discussion}, and conclude with Section~\ref{sec:conclusion}.

\section{Problem Formulation}\label{sec:problem}

In this paper, we consider rigid-body systems in the standard ``manipulator'' form:
\begin{equation}\label{eq:multibody_dynamics}
    \M(\q)\vd + \C(\q,\v)\v + \g(\q) = \S^T\btau + \sum_{c}\J_c(\q)^T\f_c
\end{equation}
where $\q$ are generalized positions (e.g. joint angles and free body poses) and $\v$ are generalized velocities (e.g. joint velocities and free body linear/angular velocities). $\M$ is the positive definite mass matrix, $\C$ and $\g$ collect Coriolis/centripetal and gravitational terms, $\btau$ are applied joint torques, $\f_c$ represents the contact wrench associated with contact $c$, and $\J_c$ is the corresponding Jacobian.

The dynamics (\ref{eq:multibody_dynamics}) describe both legged locomotion (where $\q$ consists of joint angles and a body pose) as well as manipulation (where $\q$ also includes poses of objects in the environment). We assume that $\q$ and $\v$ can be measured perfectly at any time. 

Furthermore, we assume that (\ref{eq:multibody_dynamics}) is discretized as
\begin{equation}\label{eq:discrete_dynamics}
    \x_{k+1} = f(\x_k,\u_k),
\end{equation}
where $\x_k^T = [\q_k^T~\v_k^T] \in \mathbb{R}^n$ is the system state at the $k^{th}$ time step and $\u_k = \btau_k \in \mathbb{R}^m$ are control inputs. Implicit in this formulation is the fact that contact forces $\f_c$ are computed at each time step as some function of the state $\x$ and input $\u$. Importantly, we assume that $f$ is differentiable even when making and breaking contact, i.e., 
$f_{\x} = \frac{\partial f(\x,\u)}{\partial \x}$ and $f_{\u} = \frac{\partial f(\x,\u)}{\partial \u}$ are well-defined for any $\x$ and $\u$. 

All of these assumptions are met by the open-source Drake simulator \cite{drake}, which discretizes (\ref{eq:multibody_dynamics}) using the semi-implicit integration scheme described in \cite{castro2020transition}, computes contact wrenches using the hydroelastic contact model \cite{elandt2019pressure,masterjohn2021discrete}, and makes dynamics gradients $f_{\x}$ and $f_{\u}$ available via automatic differentiation. 

With this in mind, we aim to solve contact-implicit trajectory optimization problems of the following standard form:
\begin{subequations}\label{eq:trajectory_optimization}
\begin{align}
    \min~ & \sum_{k=0}^{N-1}\left\{ \tilde{\x}_k^T\Q\tilde{\x}_k + \u_k^T\R\u_k\right\} + \tilde{\x}_N^T\Q_f\tilde{\x}_N, \\
    \mathrm{s.t.~} & \x_0 \text{ fixed}, \\
                   & \x_{k+1} = f(\x_k,\u_k),
\end{align}
\end{subequations}
where $\tilde{\x}_k = \x_k - \x^{nom}$ represents an error with respect to a nominal state $\x^{nom}$, $\Q$, $\R$, and $\Q_f$ are symmetric positive semi-definite weighting matrices, and the contact-implicit nature of the problem comes from the fact that the discretized dynamics (\ref{eq:discrete_dynamics}) account for contact interactions. 

\section{Background}\label{sec:background}
\subsection{Hydroelastic Contact}\label{sec:hydroelastic}

In this section we provide a brief overview of hydroelastic contact, also known as Pressure Field Contact (PFC). This approach was first proposed in \cite{elandt2019pressure}, with further refinements in \cite{masterjohn2021discrete}. Further details can be found in these original works, as well as in the Drake documentation \cite{drake}. 

\begin{figure}
    \centering
    \includegraphics[width=0.6\linewidth]{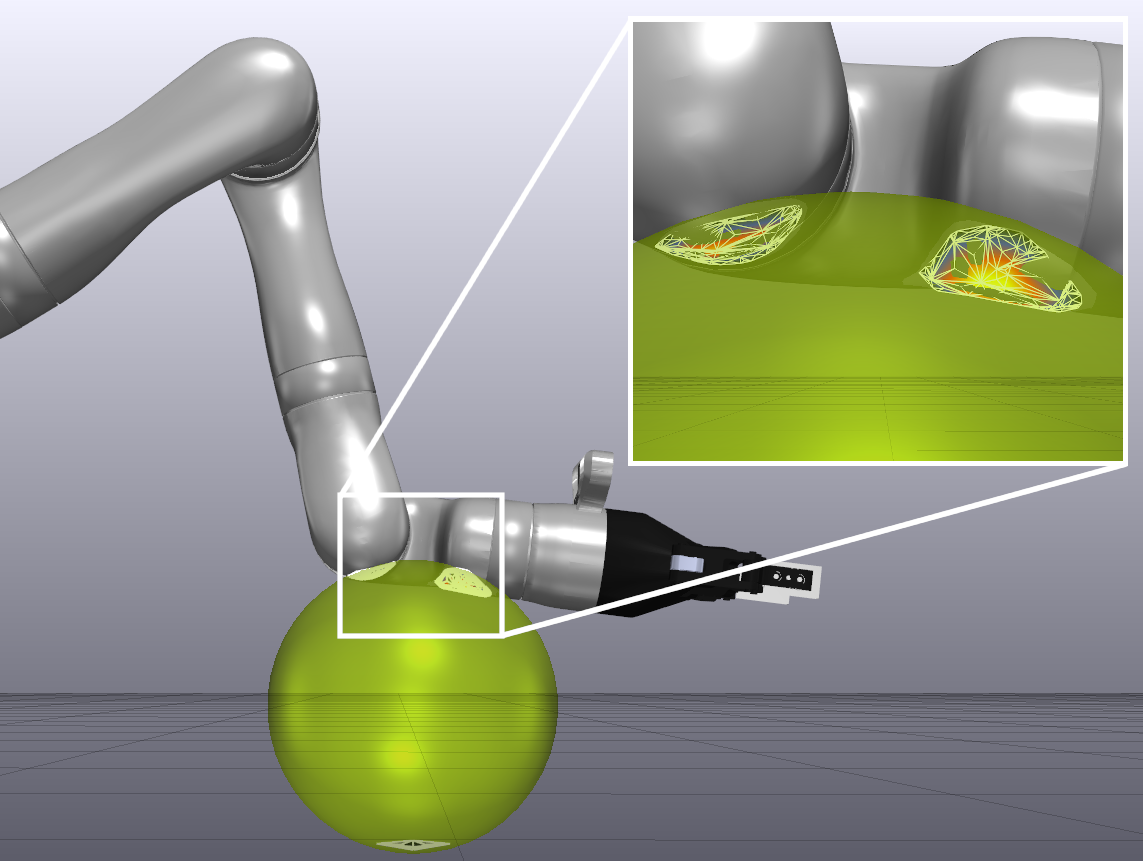}
    \caption{Visualization of hydroelastic contact surfaces between a Kinova Gen3 manipulator arm and a ball in Drake \cite{drake}.}
    \label{fig:hydroelastic}
\end{figure}

The basic idea is as follows. Discretization of the multibody dynamics (\ref{eq:multibody_dynamics}) leads to a contact-modeling problem: two bodies may be separate at time $k$, but interpenetrating at time $k+1$. The hydroelastic model resolves this problem by allowing some overlap between nominally rigid objects. Contact forces are a function of the resulting overlap, with larger overlap resulting in larger forces. 

These overlaps are resolved into contact forces by the pressure field theory described in \cite{elandt2019pressure}. Essentially, each object is associated with an internal pressure field, typically with higher pressures toward the interior of an object. When two objects overlap, the two pressure fields define an equilibrium surface. \hl{Contact impulses are computed by integrating over this surface, in conjunction with a Coulomb model of regularized friction \cite{masterjohn2021discrete}.} In practice, contact surfaces are defined by user-specified meshes. A visualization of such contact surfaces is shown in Figure~\ref{fig:hydroelastic}.

In addition to standard parameters like friction coefficients, hydroelastic contact requires the specification of \textit{hydroelastic modulus} and \textit{dissipation} parameters for each object. The hydroelastic modulus, measured in Pa ($\mathrm{N/m^2}$), defines how the pressure field increases with distance to the center of the object, with lower values corresponding to more compliant behavior. Dissipation, measured in s/m, controls how energy is lost during the contact interaction. We refer the interested reader to the Drake documentation \cite{drake} for further details. 

Two key features of hydroelastic contact make it appealing for contact-implicit trajectory optimization. The first is physical realism: contact rich interactions like that shown in Figure~\ref{fig:hydroelastic} are essential for tasks like whole-arm manipulation, but are notoriously difficult to model accurately. The second is differentiability. Drake's automatic differentiation tools enable simple computation of the dynamics partials $f_{\x}$ and $f_{\u}$ even through contact. While this is computationally expensive, it is easy to implement and provides an important proof-of-concept regarding the usefulness of the underlying gradients.

\subsection{Iterative LQR}\label{sec:ilqr}

In this section, we present a brief overview of the iLQR algorithm \cite{li2004iterative}. iLQR and its second-order variant, Differential Dynamic Programming (DDP) \cite{mayne1966second}, are popular due to rapid convergence as well as that fact that a local feedback controller is generated alongside an optimal trajectory. 

The basic idea behind iLQR is to use a linear approximation of the system dynamics and a quadratic approximation of the optimal cost-to-go at each iteration. More specifically, DDP/iLQR considers optimal control problems of the form
\begin{subequations}
\begin{align}
    \min_{\u}~ & \sum_{k=0}^{N-1}l(\x_k,\u_k) + l_f(\x_N) \\
    \mathrm{s.t.~} & \x_0 \text{ fixed} \\
                   & \x_{k+1} = f(\x_k,\u_k),
\end{align}
\end{subequations}
of which problem (\ref{eq:trajectory_optimization}) is clearly a special case. DDP/iLQR is composed of a sequence of forward and backwards passes. First, the system is simulated forward to produce a nominal trajectory with control inputs $\bar{\u}$ and states $\bar{\x}$. In the backwards pass, we consider the optimal cost-to-go $V(\x,k)$ defined by the Bellman equation:
\begin{equation*}
    V(\x,k) = \min_{\u}[l(\x,\u) + V(f(\x,\u),k+1)] = \min_{\u} Q_k(\x,\u),
\end{equation*}
where $V(\x,N) = l_f(\x)$. We then consider a second-order approximation of $Q_k$ around ($\bar{\u},\bar{\x}$):
\begin{multline*}
    \delta Q_k(\delta \x, \delta \u) = Q_k(\bar{\x}+\delta \x, \bar{\u}+\delta \u) - Q_k(\bar{\x},\bar{\u}) \\
    \approx \frac{1}{2}\begin{bmatrix}1 \\ \delta \x \\ \delta \u \end{bmatrix}^T
    \begin{bmatrix}
        0 & Q_{\x}^T & Q_{\u}^T \\ 
        Q_{\x} & Q_{\x\x} & Q_{\x\u} \\
        Q_{\u} & Q_{\u\x} & Q_{\u\u}
    \end{bmatrix}
    \begin{bmatrix}1 \\ \delta \x \\ \delta \u\end{bmatrix}.
\end{multline*}
The coefficients of this expansion can be written in terms of the cost-to-go at the following timestep, where we write $V(\x,k+1) = V'$ for conciseness:
\begin{subequations}
\begin{align}
    Q_{\x} &= l_{\x} + f_{\x}^TV_{\x}' \\
    Q_{\u} &= l_{\u} + f_{\u}^TV_{\x}' \\
    Q_{\x\x} &= l_{\x\x} + f_{\x}^TV_{\x\x}' + V_{\x}' \cdot f_{\x\x} \\
    Q_{\u\u} &= l_{\u\u} + f_{\u}^TV_{\x\x}' + V_{\x}' \cdot f_{\u\u} \\
    Q_{\u\x} &= l_{\u\x} + f_{\u}^TV_{\x\x}' + V_{\x}' \cdot f_{\u\x}
\end{align}
\end{subequations}

The terms involving second-order dynamics partials (e.g., $V_{\x}' \cdot f_{\x\x}$) are used in DDP but dropped in iLQR. Including these terms improves convergence \cite{mayne1966second}, but at the cost of additional complexity. While there are promising recent results on computing these second-order terms more efficiently \cite{nganga2021accelerating}, we focus in this paper on iLQR rather than DDP in the interest of easy implementation. 

To complete the backwards pass, $V_{\x}$ and $V_{\x\x}$ at the current timestep can be computed as
\begin{subequations}
\begin{align}
    V_{\x} &= Q_{\x} - Q_{\u}^TQ_{\u\u}^{-1}Q_{\u\x} \\
    V_{\x\x} &= Q_{\x\x} - Q_{\u\x}^TQ_{\u\u}^{-1}Q_{\u\x}.
\end{align}
\end{subequations}

The nominal trajectory is then updated in a forward pass, where new control inputs are chosen according to the control law
\begin{equation}
    \u_k = \bar{\u}_k - \epsilon \bm{\kappa}_k - \K_k(\x_k -\bar{\x}_k),
\end{equation}
where $\K_k = Q_{\u\u}^{-1}Q_{\u\x}$, $\bm{\kappa}_k = Q_{\u\u}^{-1}Q_{\u}$, and $\epsilon \in (0,1]$ is a linesearch parameter used to ensure that the cost decreases monotonically. 

The updated trajectory is used to perform a subsequent backwards pass, and the process repeats until convergence. After convergence, the local control policy
\begin{equation}\label{eq:ilqr_local_feedback}
    \u_k = \bar{\u}_k - \K_k(\x_k -\bar{\x}_k)
\end{equation}
is optimal in the neighborhood of the nominal trajectory. 

\section{Simulation Results}\label{sec:simulation}

In this section we present simulation results applying iLQR to systems with hydroelastic contact. We used a simple Python implementation of iLQR, available at \url{https://bit.ly/ilqr_hc}. We use the Drake simulator \cite{drake} and Drake's automatic differentiation tools to compute the dynamics partials $f_{\x}$ and $f_{\u}$. All experiments were performed on a laptop with an Intel i7 processor and 32GB RAM. 

\subsection{Quadruped Gait Generation}

In this section, we show how iLQR over hydroelastic contact can be used for automatic gait generation for a Mini Cheetah quadruped \cite{katz2019mini} walking over flat ground. No a-priori contact sequence or reference motions were specified, only a quadratic cost of the form (\ref{eq:trajectory_optimization}), which aims to drive the robot forward at a desired velocity. 

\begin{figure*}
    \centering
    \includegraphics[width=0.8\linewidth]{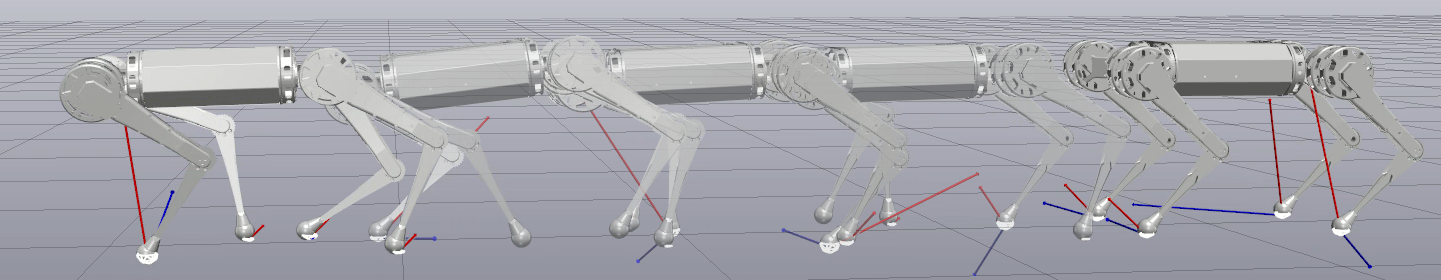}
    \caption{Simulation snapshots of optimal trajectory with a target velocity of 1 m/s. The contact sequence was not specified a priori, and the solver discovers a gait automatically. Opposite pairs of legs (e.g., front left and back right) tend to move in tandem, similar to the trotting gait exhibited by quadrupeds in nature.}
    \label{fig:quadruped_composite}
\end{figure*}

\textbf{Model Details: } The Mini Cheetah has 18 degrees of freedom, from 12 joints and the floating base. The state is 
\begin{equation}
    \x^T = [\q^T~\v^T] = [\begin{array}{ccc;{2pt/2pt}ccc}\bm{\theta}^T_B&\bm{p}^T_B&\q^T_J&\bm{\omega}^T_B&\dot{\bm{p}}^T_B&\v^T_J\end{array}],
\end{equation}
where $\bm{\theta}_B$ is the body orientation (as a quaternion), $\bm{p}_B$ is the body position, $\q_J$ are joint angles, $\bm{\omega}_B$ is the body angular velocity, $\dot{\bm{p}}_B$ is the body linear velocity, and $\v_J$ are joint velocities. The system is discretized with a 5~ms timestep.

Contact interactions were only considered between the feet, the ground, and the main body: the collision geometry of the legs was ignored for simplicity. The ground was modeled as a large (25m $\times$ 25m $\times$ 1m) box with a hydroelastic modulus of $5\times10^6$~Pa and dissipation of 0~s/m. With a coefficient of friction of $0.5$, this roughly approximates a hard floor. The feet were modeled as perfectly rigid spheres (i.e., infinite hydroelastic modulus) and the body as a compliant box with a small hydroelastic modulus of $1\times10^4$~Pa. We found that this ``soft body'' helped iLQR convergence, since some of the early trajectories involve the body striking the ground. 

\textbf{Initialization: } The robot's initial state was the stationary standing position shown at the right of Figure \ref{fig:quad_gait}. The initial guess $\bar{\u}$ was joint torques to hold this standing position. 

\hl{For long trajectories, we found that the time-stepping solver TAMSI \cite{castro2020transition} often failed to converge, leading to early termination of the optimization problem. While this problem could probably be alleviated with the use of a convex time-stepping scheme \cite{castro2021unconstrained}, we increased the trajectory length by solving} the optimization problem in receding horizon fashion. We first solved the iLQR problem with a time horizon of 0.2s, or 40 timesteps. We then shifted the horizon forward by 4 timesteps, used the optimal control sequence from the prior time window to generate a new initial guess $\bar{\u}$, and resolved the optimization problem. We repeated this process 100 times to obtain a trajectory with a total length of roughly 2 seconds.  

\textbf{Cost Function: } The cost function was designed to move the robot forward at a desired velocity, $v^{des}_x$. The nominal state $\x^{nom}$ was composed of the following elements: base orientation $\bm{\theta}_B$ corresponding to a level body, base position $\bm{p}_B$ shifted forward according to the desired velocity, joint angles $\q_J$ corresponding to the initial standing posture, base angular velocity $\bm{\omega}_B=0$, base linear velocity $\dot{\bm{p}}_B = [v^{des}_x~0~0]$, and joint velocities $\v_J = 0$.

The cost weights $\Q,\Q_f$ were defined as diagonal matrices
\begin{align*}
    \Q &=   \text{diag}([\begin{array}{ccc;{2pt/2pt}ccc} ~2.0 & 1.0 & 0.0 & 0.01 & 0.01 & 0.01 \end{array}]), \\
    \Q_f &= \text{diag}([\begin{array}{ccc;{2pt/2pt}ccc}
        \undermat{\bm{\theta}_B}{10.0} & 
        \undermat{\p_B}{5.0} & 
        \undermat{\q_J}{0.1} & 
        \undermat{\bm{\omega}_B}{1.0~} & 
        \undermat{\dot{\p}_B}{1.0~} &
        \undermat{\v_J}{0.01}\end{array}]). \\
\end{align*}
Note that the running cost $\Q$ puts no penalty on joint angles, and that the largest cost terms are related to the body position and velocity. The control penalty was $\R=0.01\bm{I}$. 

\begin{figure}
    \centering
    \includegraphics[width=0.7\linewidth]{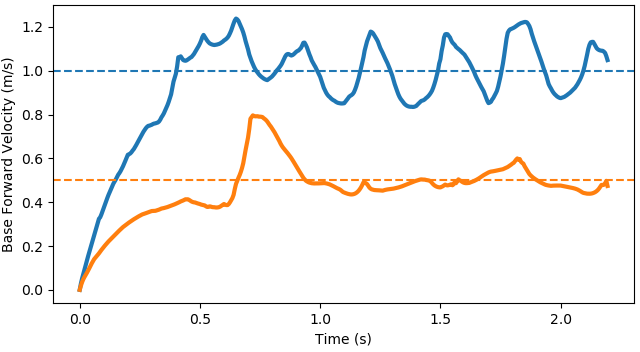}
    \caption{Actual (solid) and desired (dashed) forward base velocity of the mini cheetah quadruped for two target velocities.}
    \label{fig:forward_vel_plot}
\end{figure}

\textbf{Results: } We solved the optimization problem with target velocities of 0.5 and 1.0~m/s. Optimal trajectories can be seen in the accompanying video. Plots of the forward base velocity for both cases are shown in Figure~\ref{fig:forward_vel_plot}. Snapshots from the 1.0~m/s case are shown in Figure \ref{fig:quadruped_composite}. Interestingly, both trajectories exhibit trot-like behavior, where opposite pairs of legs tend to move together. 

Solve times were very slow in both cases, averaging around 3 seconds per iteration, as shown in Figure~\ref{fig:iteration_times}. This is largely due to the use of automatic differentiation to compute dynamics partials $f_{\x}$ and $f_{\u}$. Each receding horizon resolve required 5-6 iLQR iterations, leading to total solve times around 25 minutes. 

\begin{figure*}
    \centering
    \includegraphics[width=\linewidth]{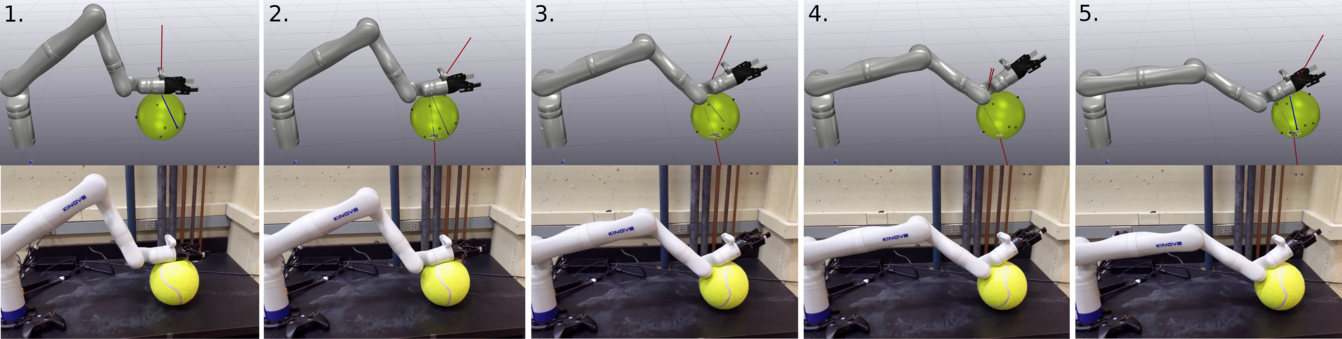}
    \caption{Snapshots of a forward push motion synthesized with iLQR over hydroelastic contact in simulation and on hardware. (1) The robot starts just above the ball and lowers its arm to make contact. (2) The robot starts to roll the ball forward. (3) As the next link of the robot strikes the ball, the ball transitions from a rolling to a sliding mode. (4) The ball continues to slide forward. (5) The arm comes to rest on top of the ball, stopping its forward motion.}
    \label{fig:forward_push_hardware}
\end{figure*}

\subsection{Whole-Arm Manipulation}\label{sec:arm}

In this section, we consider a 7 degree-of-freedom Kinova Gen3 robot arm tasked with moving a large ball. The ball is too big to be grasped, and the robot must instead exploit contact interactions with the whole arm. 

\textbf{Model Details: } The system state includes both the arm and the ball, leading to 13 total degrees of freedom. For simplicity, we assume the gripper is fixed in the open position. This gives
\begin{equation}
    \x^T = [\q^T~\v^T] = [\begin{array}{ccc;{2pt/2pt}ccc}\q_J^T&\bm{\theta}_b^T&\p_b^T&\v_J^T&\bm{\omega}_b^T&\dot{\p}_b^T\end{array}]
\end{equation}
where $\q_J$ are joint angles, $\bm{\theta}_b$ is the ball's orientation, expressed as a quaternion, $\p_b$ is the ball's position, $\v_J$ are joint velocities, $\bm{\omega}_b$ is the ball's angular velocity, and $\dot{\p}_b$ is the ball's linear velocity. A timestep of 10~ms was used to simulate the system over a 0.5~s horizon. 

Parameters of the ball were chosen to roughly match those of an oversized tennis ball used in the hardware experiments (see Section~\ref{sec:hardware}). The ball has a radius of 0.1~m, mass of 0.258~kg, coefficient of friction of $0.2$, and is modeled as a hollow sphere. We used a hydroelastic modulus of $5\times10^6$~Pa and a relatively large dissipation of 5~s/m to model the ball. The same ground model was used as in the quadruped example described above.

The collision model for the robot arm was defined using mesh files supplied by Kinova. The original high-resolution meshes were downsampled to improve the efficiency of the hydroelastic contact engine. The robot arm was modeled as perfectly rigid, i.e., infinite hydroelastic modulus. 

\textbf{Initialization} An initial guess of $\bar{\u} = \S \g$ was chosen to simply compensate for gravity. Because the time horizon was relatively short, we did not use any receding-horizon resolves. We did take care to define initial joint angles such that the robot was close to the ball.

\textbf{Cost Function: } The quadratic cost was defined to prioritize the movement of the ball while minimally restricting the robot arm itself. In addition to a control penalty $\R=0.01\bm{I}$, $\Q$ and $\Q_f$ were defined as
\begin{align*}
    \Q &= \text{diag}([\begin{array}{ccc;{2pt/2pt}ccc} 0.0 & 0.0 & 100 & 0.1 & 0.1 & 0.1 \end{array}]), \\
    \Q_f &= \text{diag}([\begin{array}{ccc;{2pt/2pt}ccc}
        \undermat{\q_J}{0.0} & 
        \undermat{\bm{\theta}_b}{0.0} & 
        \undermat{\p_b}{100} & 
        \undermat{\v_J}{0.1} & 
        \undermat{\bm{\omega}_b}{1.0} & 
        \undermat{\dot{\p}_b}{1.0}\end{array}]). \\
\end{align*}
Note that there is no cost associated with the joint angles $\q_J$, and that the primary focus is on $\p_b$.  The nominal state $\x^{nom}$ was defined by a target ball position $\p_b^{nom}$ and zero velocity.

\begin{figure}
    \centering
    \includegraphics[width=0.8\linewidth]{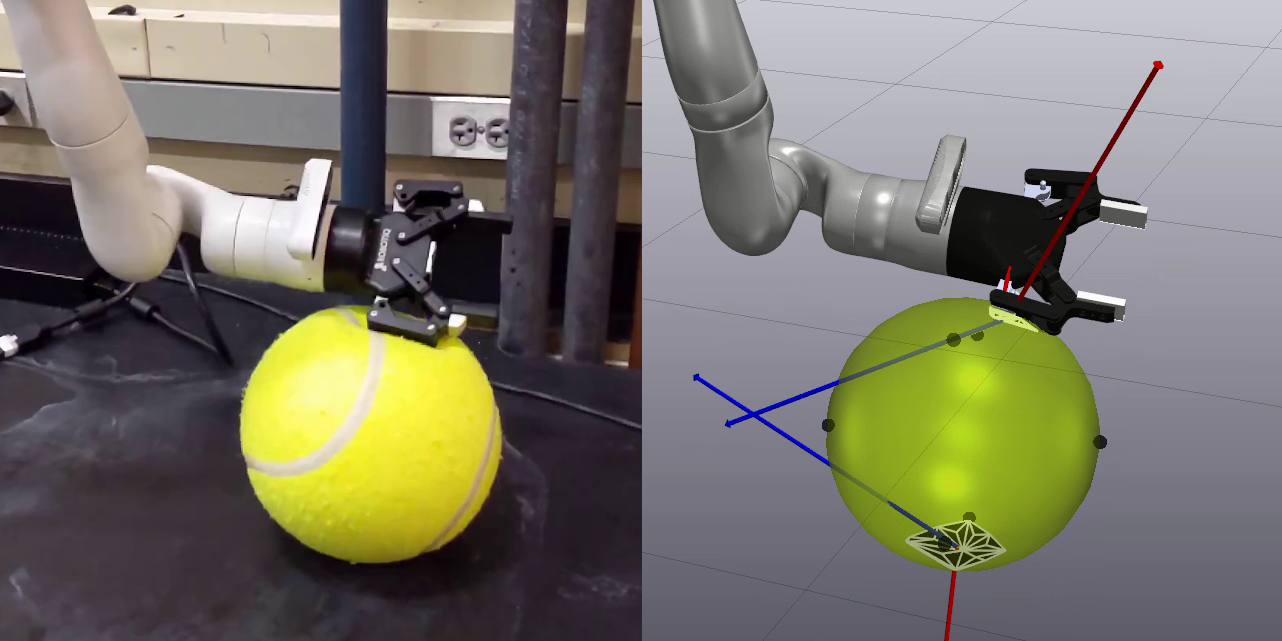}
    \caption{Comparison of actual (hardware) and planned (simulation) tests for rolling the ball to the left.}
    \label{fig:sideways_roll}
\end{figure}

\textbf{Results: } We considered three target positions $\p_b^{nom}$, defined to move the ball forward 0.2~m, left 0.15~m, and up 0.2~m. The same cost function was used in all three cases, with the exception that no penalty on the horizontal position of the ball was applied in the lifting up case. Screenshots from the resulting trajectories are shown in Figures~\ref{fig:forward_push_hardware}, \ref{fig:sideways_roll}, and \ref{fig:front_page}.

In the forward scenario, the robot uses a combination of rolling and sliding to move the ball forward. In the leftward scenario, the ball is rolled through the entire motion, with the side of the gripper used to extend the roll. In the upward lifting scenario, the robot uses contact with several of the links to roll the ball against the robot's base.

\begin{figure}
    \centering
    \includegraphics[width=0.8\linewidth]{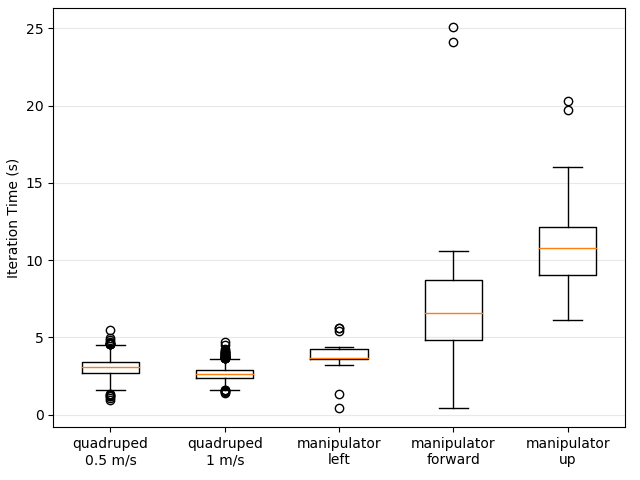}
    \caption{Box plots showing iteration times for each of the examples. }
    \label{fig:iteration_times}
\end{figure}

Iteration times for iLQR were again very slow, averaging 4-12 seconds as shown in Figure~\ref{fig:iteration_times}. The forward and upward scenarios were significantly slower due to the more complex contact interactions involved. The three scenarios (left, forward, up) required 21, 30, and 33 iterations respectively, leading to total solve times of 80, 225, and 361 seconds.

\hl{\subsection{Cart-Pole With Wall}}

\begin{figure}
    \centering
    \includegraphics[width=\linewidth]{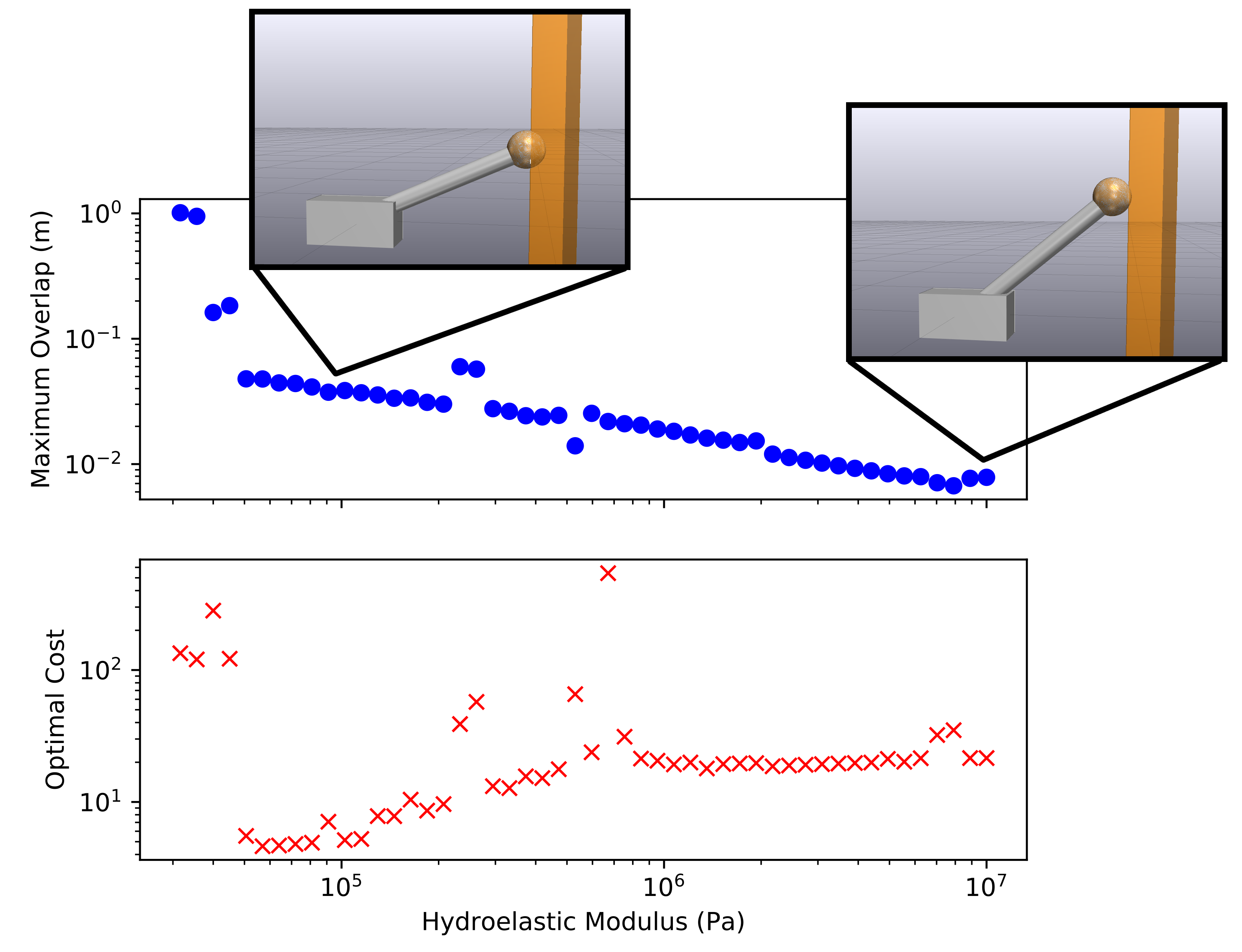}
    \caption{A cart-pole system uses a compliant wall to push itself into an upright position. Softer walls (lower hydroelastic modulus) generally result in better solutions (lower optimal cost), but only up to a certain point, after which tunnelling artifacts lead to much worse quality solutions.}
    \label{fig:cart_pole_with_wall}
\end{figure}

\hl{Hydroelastic contact can model a variety of materials, ranging from near-rigid to very compliant. In this section, we explore the effect of material softness on optimization quality using a relatively simple system with contact---a cart-pole next to a wall. This system is illustrated in Figure~\ref{fig:cart_pole_with_wall}: further details on this standard benchmark system can be found in \cite{aydinoglu2021real} and references therein.}

\hl{The objective is to balance the pole in the upright position a short distance away from the wall. Contact between the end of the pole and the wall can be used to achieve this goal. We model the wall as a rigid (infinite hydroelastic modulus) box and the end of the pole as a sphere with variable compliance. The contact interactions in this case are relatively simple, and our iLQR implementation typically converges in 1-2 seconds.}

\hl{We use iLQR to find (locally) optimal trajectories for a variety of hydroelastic moduli. The results are shown in Figure~\ref{fig:hydroelastic}, where lower hydroelastic moduli correspond to softer collisions. The top plot shows the maximum overlap between the pole and the wall over the course of each trajectory. The configurations of maximum overlap are illustrated for several representative trajectories. Maximum overlap increases fairly smoothly as hydroelastic modulus decreases, except for very soft walls, where maximum overlap increases dramatically. This is due to tunnelling artifacts, where the pole passes completely through the wall. }

\hl{The bottom plot shows the optimal cost for the same trajectories. Softer walls generally result in lower optimal costs, at least until tunnelling artifacts present an issue. This can probably be attributed to the fact that gradients through soft contact are more smooth, and lead to iLQR problems with better numerical conditioning. On the other end of the spectrum, the stiff gradients resulting from higher hydroelastic moduli may cause the optimizer to get stuck in lower quality local minima. Nonetheless, iLQR finds reasonable solutions even for very high hydroelastic moduli, which provide a close approximation of rigid contact. }

\section{Hardware Experiments}\label{sec:hardware}

To validate the physical realism of this approach, we executed the optimal trajectories generated in Section~\ref{sec:arm} on a Kinova Gen3 manipulator. 
Trajectories generated offline were executed in open-loop. Due to the fact that direct measurements of the ball's state are not available, we used a stiff PD+ controller 
\begin{equation}
    \btau = \bar{\u} - \K_p(\q^J-\bar{\q}^J) - \K_d(\v^J-\bar{\v}^J),
\end{equation}
with $\K_p = 500\bm{I}$ and $\K_d = 5\bm{I}$, rather than the local feedback controller from iLQR (\ref{eq:ilqr_local_feedback}). This torque control loop was executed at 1~kHz. To stay well below the torque and velocity limits of the robot, the trajectories were executed at half speed, with $\bar{\u},\bar{\q}^J,\bar{\v}^J$ updated every 20~ms. 

Despite these simplifications, hardware experiments exhibited remarkable similarities to the simulation. Video can be found at \url{https://youtu.be/IqxJKbM8_ms}. Snapshots from the forward scenario are shown in Figure~\ref{fig:forward_push_hardware}, where the planned (simulation) trajectory includes a transition from rolling to sliding contact around frame 3. This contact mode transition also occurs on hardware, despite a lack of direct feedback related to the ball and the slower playback speed. 

The largest difference between planned and actual trajectories can be seen in frame 5 of Figure~\ref{fig:forward_push_hardware}, where the ball ends up further forward in simulation than on hardware. This difference is most likely due to the fact that the hardware trajectory was executed at half speed, reducing the ball's forward momentum between frames 4 and 5. A lack of feedback related to the ball's position and imprecise initial placement of the ball are also possible contributing factors. 

Leftward and upward trajectories were also played back on hardware, as illustrated in Figures \ref{fig:front_page} and \ref{fig:sideways_roll}. 

\section{Discussion}\label{sec:discussion}

These preliminary results show that iLQR over hydroelastic contact is a good candidate for reliable and physically realistic contact-implicit trajectory optimization. Relatively few iterations ($<40$) were required to produce contact-rich whole-arm manipulation trajectories from scratch. Anecdotally, this method required relatively little cost-function tuning, and open-loop trajectories could be executed directly on hardware. The numerical reliability of this approach is supported by the fact that no special modifications to the contact model or optimization algorithm were required. Instead, a vanilla implementation of the standard iLQR algorithm was all that was needed to produce physically realistic contact-rich trajectories. 

Computation time is a major limitation, however. Iteration times on the order of seconds are far too slow for practical use, especially for MPC-style control. One source of slowness is computational overhead from our naive Python implementation of iLQR. The more prominent bottleneck, however, is automatic differentiation for the dynamics partials $f_{\x}$ and $f_{\u}$. This is especially notable in contact-rich configurations (Figure~\ref{fig:iteration_times}). More efficient computation of gradients through hydroelastic contact is a promising area of future research. Similar developments for the MuJoCo contact model \cite{todorov2014convex} reduced DDP iteration times to the order of milliseconds \cite{chatzinikolaidis2021trajectory}. 

Another limitation stems from the physical accuracy of the hydroelastic contact model. Since hydroelastic contact does not include force-at-a-distance, iLQR cannot direct the robot to make contact with distant objects, something that can be accomplished if force-at-a-distance is allowed \cite{onol2020tuning}.  With this in mind, iLQR over hydroelastic contact may be most useful in conjunction with a higher-level planner. Nonetheless, this method was able to synthesize complex behaviors that require making and breaking contact, such as quadruped locomotion, without a high-level planner or force-at-a-distance. 

\hl{Finally, while we focus in this paper on iLQR in particular, other optimization methods over hydroelastic contact may offer good performance as well. In particular, methods like direct collocation and multiple shooting may help alleviate some of the shortcomings of this approach via easier specification of initial guesses or more efficient search through the cost landscape. Whether this is indeed the case, or whether other optimization paradigms might offer further improvements, is an important area for future research.}

\section{Conclusion}\label{sec:conclusion}

iLQR over hydroelastic contact enables numerically reliable and physically realistic contact-implicit trajectory optimization. While computation time is currently a major drawback, there is good reason to believe that this is not a fundamental limitation, and that future research on computing gradients through hydroelastic contact could enable a major step towards fast and reliable contact-implicit trajectory optimization. 

\section{Acknowledgements}

Thanks to Patrick Wensing and He Li for many helpful discussions. 

\bibliographystyle{IEEEtran}
\bibliography{references}

\end{document}